# A Novel Downsampling Strategy Based on Information Complementarity for Medical Image Segmentation

Wenbo Yue⋆  Chang Li†  Guoping Xu⋆

⋆ School of Computer Science and Engineering,Wuhan Institute of Technology,Wuhan, China
†Department of Biomedical Engineering,Hefei University of Technology,Wuhan, China

*Abstract*—In convolutional neural networks (CNNs), downsampling operations are crucial to model performance. Although traditional downsampling methods (such as maximum pooling and cross-row convolution) perform well in feature aggregation, receptive field expansion, and computational reduction, they may lead to the loss of key spatial information in semantic segmentation tasks, thereby affecting the pixel-by-pixel prediction accuracy.To this end, this study proposes a downsampling method based on information complementarity - Hybrid Pooling Downsampling (HPD). The core is to replace the traditional method with MinMaxPooling, and effectively retain the light and dark contrast and detail features of the image by extracting the maximum value information of the local area.Experiment on various CNN architectures on the ACDC and Synapse datasets show that HPD outperforms traditional methods in segmentation performance, and increases the DSC coefficient by 0.5% on average. The results show that the HPD module provides an efficient solution for semantic segmentation tasks.

## I. INTRODUCTION

Convolutional Neural Networks (CNNs) play a central role in fields such as computer vision and image processing, achieving significant results in tasks like image recognition [1], object detection [2], and semantic segmentation [3]. With the development of deep convolutional neural networks (DCNNs), representative architectures like AlexNet [4], GoogLeNet [5], and ResNet [6] have emerged, driving advancements in semantic segmentation technology. In these networks, downsampling operations are crucial, commonly utilizing methods such as max pooling, average pooling, and strided convolutions to control input resolution, reduce computational complexity, and expand the receptive field. However, traditional downsampling methods have serious drawbacks, such as the loss of key information like boundaries and textures, especially when dealing with small-scale objects or complex backgrounds, where this information is essential for precise segmentation.

To mitigate this issue, researchers have proposed various improvement methods, mainly focusing on the following four directions:

First, feature transmission and fusion: By using skip connections or specific mapping rules, the feature maps downsampled by the encoder are associated with the decoder layers, enabling efficient flow and integration of information. Examples include U-Net [7], LCU-Net [8], CENet [9], LinkNet [10], and RefineNet [11]. For instance, LCU-Net adopts a hierarchical cascading mechanism, assigning downsampled features to corresponding decoder layers based on importance and using adaptive weight fusion to reduce information redundancy, improving detail reconstruction and semantic understanding.

Second, multi-scale feature extraction and fusion: By capturing and fusing features at different resolutions, the ability to segment objects at various scales is enhanced. Typical networks include DeepLab [12], PSPNet [13], PCPLP-Net [14], HRNet [15], and ICNet [16]. For example, HRNet processes multi-resolution branches in parallel, retaining low-level details and high-level semantic information, and dynamically integrates multi-scale features with weighted fusion, significantly improving segmentation accuracy for small-scale objects.

Third, multi-modal image fusion: By combining data characteristics from different modalities, complementary information is exploited to make up for the shortcomings of a single modality. Examples include DiSegNet [17], MMADT [18], CANet [19], and RGBD-Net [20]. For instance, RGBD-Net fuses RGB and depth images, using a cross-modal attention mechanism to explore the correlation between the two, improving segmentation accuracy through complementary mechanisms.

Fourth, introduction of prior information: For example, networks based on semantic boundary priors [21] use edge detection algorithms [22] to extract potential semantic boundary information and encode it as prior maps, which guide the model during downsampling and feature extraction to enhance segmentation ability in complex scenes.

However, although these methods have alleviated the information loss problem caused by traditional downsampling to a certain extent, they all have their own bottlenecks. For example, feature transfer and fusion methods may face noise interference and information attenuation during information transfer[23]; multi-scale feature extraction and fusion may affect the training and inference speed of the model due to high computational complexity; multimodal image fusion may be affected by the quality and alignment of different modal data; the introduction of prior information may mislead the learning of the model due to inaccurate prior information. Therefore, the information loss problem caused by traditional downsampling, especially the loss of key information such as boundaries, scales and textures, is still a key problem that needs to be solved in the current semantic segmentation field. Therefore, the solution we came up with is to design a downsampling method based on information complementarity, so that it can retain as much information as possible for semantic segmentation during the downsampling process, and solve this problem from the source of downsampling. Inspired by the theory of multimodal information synergy and complementarity, we propose a downsampling strategy based on information complementarity - Hybrid Pooling Downsampling (HPD).

The core process of HPD consists of two parts: first, the minmaxpooling mechanism is used to process the feature map, and its resolution and number of channels are reasonably adjusted while maintaining rich information; then, with the help of the designed convolution operation and feature screening process, irrelevant redundant information is further filtered out to ensure that the retained features are of key value to the segmentation task, laying a solid foundation for the model's excellent performance in semantic segmentation tasks.Figure 1 illustrates four types of pooling methods.

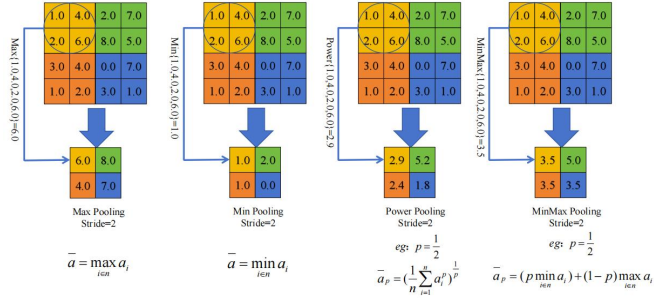

Fig.1. Illustration of four pooling methods

Figure 2 shows an example of downsampling using max pooling and minmax pooling on the basic UNet model. We can see that compared with the traditional downsampling method such as max pooling, minmaxpooling can retain more detail information and the boundaries can be observed more clearly. In summary, the main contributions of this paper are as follows: (1) We propose a downsampling strategy based on information complementarity. By fusing multiple information, we try to retain as much key information as possible during the downsampling process, thereby exploring feasible ways to reduce information loss. (2) This information complementarity strategy can directly replace the traditional downsampling method without significantly increasing the computational cost, and can be easily integrated into the existing segmentation network architecture. A large number of experimental results show that this method has shown significant effectiveness among the six most advanced SOTA segmentation methods.

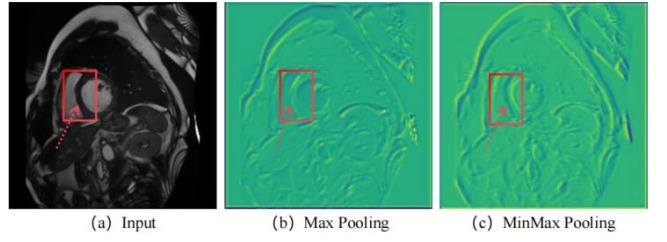

Fig.2. Downsampling examples of max pooling and minmax pooling in UNet.

## II. METHODS

As shown in Figure 3,the proposed Hybrid Pooling Downsampling (HPD) module consists of two parts: (1) feature encoding module; (2) feature learning module. The feature encoding module is responsible for transforming features and reducing spatial resolution. To achieve this goal, we use hybrid pooling transformation, which can effectively reduce the resolution of feature maps while retaining all information. The feature learning module contains a standard convolutional layer, batch normalization layer, and rectified linear unit (ReLU) activation layer, which is used to extract discriminative features. Each module will be elaborated in the following subsections.

### A. Feature Encoding Module

The feature encoding module uses a downsampling layer based on information complementarity to effectively adjust the spatial resolution of the feature map while retaining key information. HPD effectively integrates the extreme value features of the local area by extracting the minimum and maximum information of the local area, thereby more comprehensively retaining the light and dark contrast and detail features of the image while reducing the resolution of the feature map.

Specifically, for an input feature map $Z$ of dimension $H \times W \times C$ (where $H$ is height, $W$ is width, and $C$ is the number of channels), in the window division stage, assuming that the feature map is divided into windows of size (a positive integer), then in the horizontal direction, the number of windows divided is $\frac{H}{k}$, and the number of windows divided in the vertical direction is $\frac{W}{k}$, for a total of $n = \frac{H}{k} \times \frac{W}{k}$ windows.

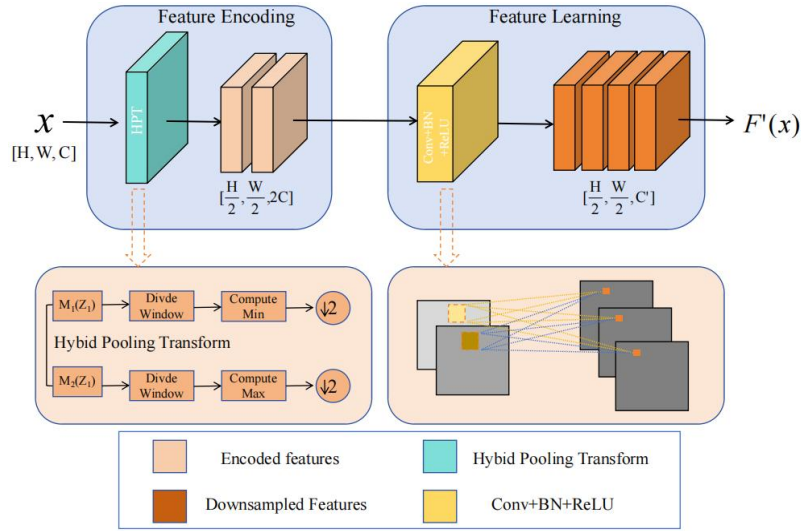

Fig.3. The architecture of the proposed HPD module

For the $i$ window, the calculation formulas for the minimum pooling value $M_1(Z_i)$ and the maximum pooling value $M_2(Z_i)$ of $Z_i (1 \leq i \leq n)$ are:

$$M_1(Z_i) = \min_{x=1, y=1}^{k,k} Z_i(x, y) \quad (1)$$

$$M_2(Z_i) = \max_{x=1, y=1}^{k,k} Z_i(x, y) \quad (2)$$

Where $Z_i(x, y)$ represents the element value with coordinates $(x, y)$ in window $Z_i$.

The new fusion feature value is shown in the following formula (3):

$$F_{\min\max}(Z_i) = M_1(Z_i) + M_2(Z_i) \quad (3)$$

In the minmaxpooling downsampling operation, the input feature map (original image size is $H \times W$) is converted into a new feature representation. After window division and feature value calculation (minmum pooling value plus maximum pooling value) and then downsampling, the downsampling operation makes the spatial resolution of each component half of the original signal, that is, $H \times W$ changes from $\frac{H}{2} \times \frac{W}{2}$. Figure 4 shows the process of minmax pooling decomposing an image with a resolution of $H \times W$. Here, the symbol $\downarrow 2$ indicates that the approximation and detail components are first downsampled. When minmaxpooling is applied to a two-dimensional signal (such as a grayscale image), it produces new feature components. In this process, minimum pooling is similar to a low-pass filter, which is used to capture the overall feature information of the local area, while maximum pooling is similar to a high-pass filter[24], which is used to capture the salient feature information of the local area. This downsampling operation can retain the key feature information in the original image while reducing the spatial resolution of the feature map, so that the subsequent network layers can extract more representative features from these transformed feature components. In addition, the number of channels of the feature map will change after downsampling, which can re-encode the information of some spatial dimensions in a way that is conducive to subsequent processing.

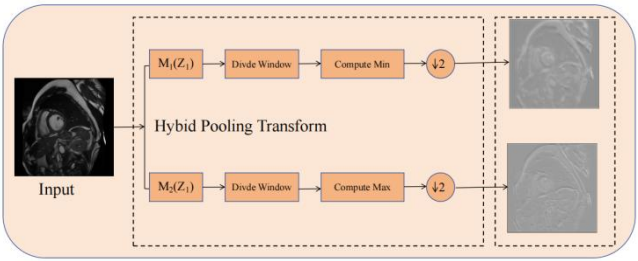

Fig.4. The illustration of the Hybid Pooling Transform.

### B. Feature Learning Module

The feature representation learning module consists of a standard $1 \times 1$ convolution layer, a batch normalization layer, and a ReLU activation function. In this module, standard convolution is used to adjust the number of channels of the feature map. This module has two main functions: (1) adjust the number of channels of the feature map to align with the subsequent layers; (2) filter redundant information as much as possible so that the subsequent layers can learn representative features more effectively. In Figure 5, we compare the output of the proposed HPD module and the maximum pooling operation. It can be seen that the output feature map of HPD has more details compared with the maximum pooling.

In general, the proposed information complementarity-based downsampling method contains two key parts. The first is to use the HPD method to reduce the spatial resolution of the feature map; the second is to use standard $1 \times 1$ convolution, batch normalization, and ReLU operations to filter redundant information.

## III. EXPERIMENTS

To evaluate the effectiveness of our HPD approach, we conduct extensive experiments on two public datasets. In this section, we briefly introduce these datasets and discuss the implementation details. Subsequently, we report the segmentation results and compare with the state-of-the-art (SOTA) methods on these two datasets.

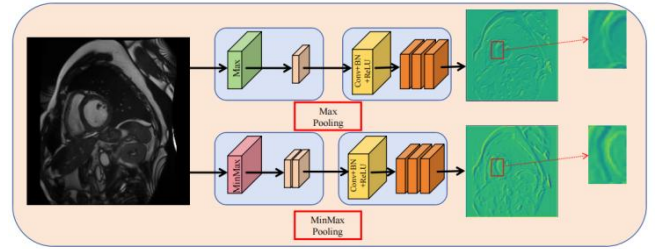

Fig.5. The illustration of HPD and max pooling downsampling methods for a MRI image.

### A. Dataset

ACDC dataset: This dataset is an important dataset for cardiac research in the field of medical imaging. It contains 1792 images and is divided into a training set (1312 images), a validation set (101 images), and a test set (380 images). In terms of performance evaluation, the focus is mainly on three key parts of the heart, namely the left ventricle (LV), the right ventricle (RV), and the myocardium, and the mean Dice similarity coefficient (mDSC) is used as the core evaluation indicator.

Synapse dataset: This dataset contains 30 abdominal CT scans, with a total of 3779 axial enhanced abdominal clinical CT images. According to the established data segmentation strategy, 18 cases were selected for training, 6 cases for validation, and the remaining 6 cases for testing. The performance evaluation is mainly based on eight important abdominal organs, namely the aorta, gallbladder (Gall), spleen, left kidney (KidL), right kidney (KidR), liver, pancreas, and stomach (Stom), and the mean Dice similarity coefficient (mDSC) is used to measure the performance of the model on this dataset.

TABLE I
SAMPLE NUMBERS FOR SEGMENTATION TASKS IN ACDC AND SYNAPSE DATASETS

| Name | Train | Validation | Test |
|---|---|---|---|
| ACDC | 1312 | 380 | 210 |
| Synapse | 2211 | 764 | 804 |

TABLE II
TRAINING PARAMETER SETTING INFORMATION FOR ACDC AND SYNAPSE

| Name | Train | Test | Val | optimizer | Lr | Batch_size | size |
|---|---|---|---|---|---|---|---|
| ACDC | 70 | 20 | 10 | SGD | 0.01 | 12 | 256 |
| Synapse | 18 | 6 | 6 | SGD | 0.01 | 12 | 256 |

### B. Implementation details

All experiments were performed with Python 3.8 and PyTorch 2.1.0, running on NVIDIA GeForce RTX 4090 (24GB video memory) and Ubuntu 18 operating system, with CUDA version 11.8. The model was trained using the SGD optimizer with an initial learning rate of 0.01 and a weight decay of $1\times10^{-4}$. The "poly" learning rate strategy was also adopted with a power of 0.9. The batch size for each iteration was 12. According to the method of Chen et al. [25], all 3D volume datasets were trained slice by slice for evaluation on the ACDC and Synapse datasets. The basic information of the overall dataset is shown in Table 1 and Table 2. The Dice similarity coefficient (DSC) is used as the evaluation metric, which is defined as follows:

$$DSC = \frac{2TP}{2TP + FP + FN} \quad (4)$$

In $TP$、$FP$ and $FN$ represent true positive examples, false positive examples and false negative examples respectively. The word "data" is plural, not singular.

### C. Experimental results on ACDC dataset

- Comparison with SOTA methods

We evaluate our method on six state-of-the-art segmentation architectures, including U-Net, LinkNet, Fast-SCNN[26], DeepLabv3+[27], Swin-Unet[28], and TransUNet[29].

Table 4 shows the Dice Similarity Coefficient (mDSC) performance comparison before and after applying Hybrid Pooling Downsampling (HPD) to replace the traditional downsampling method in 6 current best segmentation architectures. In general, the mean Dice coefficient (mDSC) of most architectures has improved after incorporating HPD. For example, the mDSC of UNet increased from 0.8993 to 0.9032 (about 0.39%), LinkNet increased from 0.869 to 0.8767 (about 0.88%), and TransUNet increased from 0.8842 to 0.8920 (about 0.78%). These results verify the effectiveness of HPD based on information complementarity design. However, for the Deeplabv3+ model, the introduction of HPD failed to improve performance, but instead caused the mDSC to drop from 0.8001 to 0.7901 (about 0.25%). After analysis, Deeplabv3+ already contains an information complementation mechanism similar to HPD. Adding HPD may introduce redundancy, which in turn leads to performance degradation. Overall, HPD significantly improves segmentation performance on multiple models, demonstrating its potential in enhancing downsampling quality.

As can be seen from Table 3, it presents the parameters and FLOPs of the five models before and after the HPD concept is incorporated. After comparative analysis, it is found that after the introduction of HPD, the parameters and FLOPs of these models have not changed significantly, which further strongly confirms the feasibility of the HPD-based method in practical applications, indicating that it will not hinder the application of the model due to a significant increase in computing resource requirements, and provides important basis and support for the promotion and use of this method in related fields.

Figure 5 shows the visualization of segmentation results based on six different models (UNet, LinkNet, FastSCNN, Swin_UNet, Trans_UNet and DeepLabv3+), and compares two different downsampling strategies (Original and MinMaxPool). The experiment is based on the ACDC dataset and focuses on analyzing the segmentation performance of the three main anatomical targets of the heart (left ventricle, right ventricle and myocardium). The following two key conclusions are summarized: (1) Improvement of segmentation accuracy: The downsampling method designed with the HPD idea shows higher accuracy in segmenting various anatomical structures of the heart, especially in the boundary area of the left ventricle and right ventricle. The prediction results are more consistent with the real annotations and are significantly better than the traditional downsampling strategy. (2) Optimization of boundary and small target details: Compared with the original downsampling, the MinMaxPool strategy has a stronger ability to capture boundary details, making the segmentation results smoother in retaining the details of the structure contour and small targets (such as the myocardium), avoiding the common boundary blur and segmentation discontinuity problems in the traditional downsampling method. The numbers in each figure represent the results of DSC.

TABLE III
PARAMETERS AND FLOPS OF FIVE MODELS BEFORE AND AFTER APPLYING HPD.

| Model | Parameters/M | FLOPs/G |
|---|---|---|
| UNet | 14.79 | 31.05 |
| UNet_MinMax | 18.70 | 35.27 |
| LinkNet | 11.53 | 3.05 |
| LinkNet_MinMax | 11.64 | 3.50 |
| Fast-SCNN | 1.14 | 0.22 |
| Fast-SCNN_MinMax | 1.15 | 0.39 |
| Swin-Unet | 27.15 | 5.93 |
| Swin-Unet_MinMax | 27.32 | 6.32 |
| TransUNet | 105.32 | 32.26 |
| TransUNet_MinMax | 106.50 | 32.56 |

- Ablation study

In order to further explore the impact of the number of minmaxpooling in the HPD module on the model performance in the UNet network architecture, this ablation experiment was conducted on the ACDC dataset. During the experiment, the basic training parameters were kept consistent with the previous ones, only the number of training rounds (epochs) was set to 300, and the Dice similarity coefficient (DSC) was used as the evaluation indicator. As shown in Table 5, different numbers of minmaxpooling have different degrees of impact on the UNet model, but overall, a reasonable configuration of the number of HPD modules can improve the model segmentation performance, which also indirectly reflects the effectiveness of the downsampling method based on information complementarity.

TABLE IV
PERFORMANCE OF SIX SOTA SEGMENTATION ARCHITECTURE ON ACDC DATASET.

| Model | mDSC | RV | MYO | LV |
|---|---|---|---|---|
| UNet | 0.8993 | **0.8958** | 0.8649 | 0.9373 |
| UNet_MinMax | **0.9032** | 0.8922 | **0.8768** | **0.9407** |
| LinkNet | 0.869 | 0.86 | 0.8214 | 0.9257 |
| LinkNet_MinMax | **0.8767** | **0.8628** | **0.8341** | **0.9331** |
| Fast-SCNN | 0.7753 | 0.7587 | 0.6967 | 0.8704 |
| Fast-SCNN_MinMax | **0.7863** | **0.7665** | **0.7086** | **0.8837** |
| Deeplabv3+ | **0.8001** | **0.7948** | **0.7633** | **0.8423** |
| Deeplabv3+_MinMax | 0.7901 | 0.7712 | 0.7578 | 0.8412 |
| Swin-Unet | 0.8721 | 0.8781 | 0.8458 | 0.8924 |
| Swin-Unet_MinMax | **0.8763** | **0.8802** | **0.8475** | **0.9003** |
| TransUNet | 0.8842 | **0.8778** | **0.8495** | 0.9053 |
| TransUNet_MinMax | **0.8920** | 0.8706 | 0.8487 | **0.9265** |

Note: The best result is marked in bold

TABLE V
THE AVERAGE PERFORMANCE OF UNET WITH DIFFERENT NUMBERS OF MINMAXPOOLING ON THE ACDC DATASET

| No.of HPD | mDSC | RV | MYO | LV |
|---|---|---|---|---|
| 0 | 0.8944 | 0.8689 | **0.8683** | 0.946 |
| 1 | 0.8949 | 0.8717 | 0.8676 | 0.9454 |
| 2 | 0.8923 | 0.8702 | 0.8647 | 0.9419 |
| 3 | **0.8975** | **0.8794** | 0.8681 | 0.9450 |
| 4 | 0.8905 | 0.8691 | 0.8619 | 0.9405 |

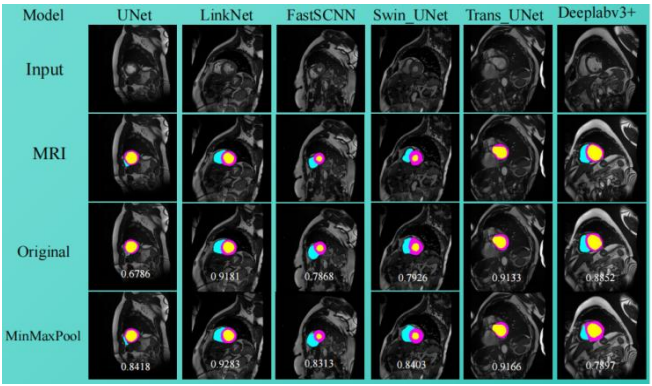

Fig.6. Visualized segmentation results on ACDC dataset.

### D. Experimental results on Synapse dataset

We selected 5 latest SOTA models and conducted experiments on the Synapse dataset to compare the effects of traditional downsampling methods and HPD methods. As shown in Table 6, after using HPD, the segmentation performance of each model on different anatomical structures was improved, verifying the effectiveness of the HPD method. For example, the mDSC of UNet increased from 0.7596 to 0.7638 (about 0.55%), and the pancreas index increased from 0.5267 to 0.5929 (6.62%). The mDSC of LinkNet increased from 0.7474 to 0.7578 (about 1.39%), and its liver index increased from 0.8843 to 0.9323.In addition, the mDSC of Swin-Unet increased from 0.6227 to 0.6531 (4.88%), and the segmentation performance of multiple anatomical structures was improved to varying degrees. The mDSC of Fast-SCNN slightly decreased from 0.6487 to 0.6465, but some organs (such as the aorta) still improved to a certain extent. Overall, the experimental results fully demonstrate that the HPD method can effectively improve the segmentation performance of the model, especially in small object and detail segmentation tasks.

## IV. CONCLUSION

This paper proposes a downsampling strategy based on information complementarity, Hybrid Pooling Downsampling (HPD). By introducing the fusion of different sampling methods, the performance of the semantic segmentation model is significantly improved while maintaining a low computational overhead. Experimental results show that HPD has significantly improved the segmentation accuracy on the ACDC and Synapse datasets, especially in the optimization of boundary details. Although HPD has shown great potential for performance improvement, its adaptability to complex scenes and large-scale datasets still needs to be further optimized. Future work will focus on exploring the lightweight design of HPD and its possible application in other task scenarios. Overall, the HPD module provides an innovative solution for the downsampling operation in semantic segmentation, and opens up a new direction for solving the problem of information loss in deep learning, showing broad application prospects and research value.

TABLE VI
PERFORMANCE OF FIVE SOTA MODELS WITH HPD ON THE SYNAPSE DATASETS

| Model | mDSC | Aorta | Gall | KidL | KidR | Liver | Pancreas | Spleen | Stomach |
|---|---|---|---|---|---|---|---|---|---|
| UNet | 0.7596 | **0.8766** | 0.6231 | 0.7245 | **0.675** | 0.9579 | 0.5267 | 0.8538 | **0.7562** |
| UNet_MinMax | **0.7638** | 0.8542 | **0.6374** | **0.7459** | 0.6374 | **0.9629** | **0.5929** | **0.8542** | 0.7542 |
| LinkNet | 0.7474 | 0.73 | **0.6613** | 0.7584 | 0.6789 | 0.8843 | 0.5432 | 0.854 | 0.7596 |
| LinkNet_MinMax | **0.7578** | **0.7441** | 0.6466 | **0.7601** | **0.6837** | **0.9323** | **0.5758** | **0.905** | **0.7684** |
| Fast-SCNN | **0.6487** | 0.6516 | 0.5636 | 0.6374 | **0.5736** | 0.8547 | **0.5063** | 0.7114 | **0.7052** |
| Fast-SCNN_MinMax | 0.6465 | 0.6504 | **0.5684** | **0.6427** | 0.5705 | **0.8635** | 0.4538 | **0.7196** | 0.7034 |
| Swin-Unet | 0.6227 | 0.6328 | 0.5263 | 0.5821 | 0.6034 | 0.8683 | **0.3684** | 0.7139 | 0.6482 |
| Swin-Unet_MinMax | **0.6531** | **0.6512** | **0.5626** | **0.6531** | **0.6453** | **0.9032** | 0.3632 | **0.7436** | **0.7048** |
| TransUNet | 0.7518 | **0.8358** | 0.5821 | 0.7538 | **0.7393** | 0.9298 | 0.5418 | **0.8399** | **0.7318** |
| TransUNet_MinMax | **0.7535** | 0.8265 | **0.6163** | **0.7621** | 0.7335 | **0.9431** | **0.5736** | 0.8378 | 0.7317 |

## REFERENCES

[1] Keysers D, Deselaers T, Gollan C, et al. Deformation models for image recognition[J]. IEEE Transactions on Pattern Analysis and Machine Intelligence, 2007, 29(8): 1422-1435.

[2] Zhao Z Q, Zheng P, Xu S, et al. Object detection with deep learning: A review[J]. IEEE transactions on neural networks and learning systems, 2019, 30(11): 3212-3232.

[3] Mo Y, Wu Y, Yang X, et al. Review the state-of-the-art technologies of semantic segmentation based on deep learning[J]. Neurocomputing, 2022, 493: 626-646.

[4] Krizhevsky A, Sutskever I, Hinton G E. Imagenet classification with deep convolutional neural networks[J]. Advances in neural information processing systems, 2012, 25.

[5] Szegedy C, Liu W, Jia Y, et al. Going deeper with convolutions[C]//Proceedings of the IEEE conference on computer vision and pattern recognition. 2015: 1-9.

[6] He K, Zhang X, Ren S, et al. Deep residual learning for image recognition[C]//Proceedings of the IEEE conference on computer vision and pattern recognition. 2016: 770-778.

[7] Ronneberger O, Fischer P, Brox T. U-net: Convolutional networks for biomedical image segmentation[C]//Medical image computing and computer-assisted intervention–MICCAI 2015: 18th international conference, Munich, Germany, October 5-9, 2015, proceedings, part III 18. Springer International Publishing, 2015: 234-241.\

[8] Zhang J, Li C, Kosov S, et al. LCU-Net: A novel low-cost U-Net for environmental microorganism image segmentation[J]. Pattern Recognition, 2021, 115: 107885.

[9] Zhou Q, Wu X, Zhang S, et al. Contextual ensemble network for semantic segmentation[J]. Pattern Recognition, 2022, 122: 108290.

[10] Chaurasia A, Culurciello E. Linknet: Exploiting encoder representations for efficient semantic segmentation[C]//2017 IEEE visual communications and image processing (VCIP). IEEE, 2017: 1-4.

[11] Lin G, Milan A, Shen C, et al. Refinenet: Multi-path refinement networks for high-resolution semantic segmentation[C]//Proceedings of the IEEE conference on computer vision and pattern recognition. 2017: 1925-1934.

[12] Chen L C, Zhu Y, Papandreou G, et al. Encoder-decoder with atrous separable convolution for semantic image segmentation[C]//Proceedings of the European conference on computer vision (ECCV). 2018: 801-818.

[13] Zhao H, Shi J, Qi X, et al. Pyramid scene parsing network[C]//Proceedings of the IEEE conference on computer vision and pattern recognition. 2017: 2881-2890.

[14] Mu N, Wang H, Zhang Y, et al. Progressive global perception and local polishing network for lung infection segmentation of COVID-19 CT images[J]. Pattern Recognition, 2021, 120: 108168.

[15] Wang J, Sun K, Cheng T, et al. Deep high-resolution representation learning for visual recognition[J]. IEEE transactions on pattern analysis and machine intelligence, 2020, 43(10): 3349-3364.

[16] Zhao H, Qi X, Shen X, et al. Icnet for real-time semantic segmentation on high-resolution images[C]//Proceedings of the European conference on computer vision (ECCV). 2018: 405-420.

[17] Xu G, Cao H, Udupa J K, et al. DiSegNet: A deep dilated convolutional encoder-decoder architecture for lymph node segmentation on PET/CT images[J]. Computerized Medical Imaging and Graphics, 2021, 88: 101851.

[18] Hu S, Bonardi F, Bouchafa S, et al. Multi-modal unsupervised domain adaptation for semantic image segmentation[J]. Pattern Recognition, 2023, 137: 109299.

[19] Zhou H, Qi L, Huang H, et al. CANet: Co-attention network for RGB-D semantic segmentation[J]. Pattern Recognition, 2022, 124: 108468.

[20] Nguyen P, Karnewar A, Huynh L, et al. Rgbd-net: Predicting color and depth images for novel views synthesis[C]//2021 International Conference on 3D Vision (3DV). IEEE, 2021: 1095-1105.

[21] Bertasius G, Shi J, Torresani L. Semantic segmentation with boundary neural fields[C]//Proceedings of the IEEE conference on computer vision and pattern recognition. 2016: 3602-3610.

[22] Ziou D, Tabbone S. Edge detection techniques-an overview[J]. Распознавание образов и анализ изображен/Pattern Recognition and Image Analysis: Advances in Mathematical Theory and Applications, 1998, 8(4): 537-559.

[23] Cossu R, Chiappini L. A color image segmentation method as used in the study of ancient monument decay[J]. Journal of Cultural Heritage, 2004, 5(4): 385-391.

[24] Wu D, Fang N, Sun C, et al. Terahertz plasmonic high pass filter[J]. Applied Physics Letters, 2003, 83(1): 201-203.

[25] X. Li, M. He, H. Li, H. Shen, A combined loss-based multiscale fully convolutional network for high-resolution remote sensing image change detection, IEEE Geosci.Remote Sens. Lett. 19 (2021) 1–5.

[26] Poudel R P K, Liwicki S, Cipolla R. Fast-scnn: Fast semantic segmentation network[J]. arXiv preprint arXiv:1902.04502, 2019.

[27] Chen L C. Rethinking atrous convolution for semantic image segmentation[J]. arXiv preprint arXiv:1706.05587, 2017.

[28] Cao H, Wang Y, Chen J, et al. Swin-unet: Unet-like pure transformer for medical image segmentation[C]//European conference on computer vision. Cham: Springer Nature Switzerland, 2022: 205-218.

[29] Chen J, Lu Y, Yu Q, et al. Transunet: Transformers make strong encoders for medical image segmentation[J]. arXiv preprint arXiv:2102.04306, 2021.